\documentclass[conference]{IEEEtran}
\usepackage{cite}
\usepackage{amsmath,amssymb,amsfonts}
\usepackage{algorithmic}
\usepackage{graphicx}
\usepackage{textcomp}
\usepackage{xcolor}
\usepackage{booktabs}
\usepackage{multirow} 
\usepackage{url} 
\usepackage[table]{xcolor}
   
\def\BibTeX{{\rm B\kern-.05em{\sc i\kern-.025em b}\kern-.08em
    T\kern-.1667em\lower.7ex\hbox{E}\kern-.125emX}}

\makeatletter
\newcommand{\maketitlesupplementary}{%
   \newpage
   \twocolumn[%
        \centering
        \Large
        \textbf{\@title}\\
        \vspace{0.5em}%
        Supplementary Material \\%
        \vspace{1.0em}%
   ]%
}
\makeatother

\begin{document}

\title{Uni-Classifier: Leveraging Video Diffusion Priors for Universal Guidance Classifier}

\author{\textbf{Yujie Zhou}$^{1*}$\quad
\textbf{Pengyang Ling}$^{2*}$\quad
\textbf{Jiazi Bu}$^{1*}$\quad
\textbf{Bingjie Gao}$^{1}$\quad
\textbf{Li Niu}$^{1,3\dag}$ \\
\textsuperscript{\rm 1}Shanghai Jiao Tong University \
\textsuperscript{\rm 2}University of Science and Technology of China \
\textsuperscript{\rm 3}Miguo.ai \\
} 

\maketitle

{
  \renewcommand{\thefootnote}{\fnsymbol{footnote}}
  \footnotetext[1]{Equal contribution. \textsuperscript{\dag} Corresponding authors.}
}

\begin{abstract}
In practical AI workflows, complex tasks often involve chaining multiple generative models, such as using a video or 3D generation model after a 2D image generator. However, distributional mismatches between the output of upstream models and the expected input of downstream models frequently degrade overall generation quality. To address this issue, we propose Uni-Classifier (Uni-C), a simple yet effective plug-and-play module that leverages video diffusion priors to guide the denoising process of preceding models, thereby aligning their outputs with downstream requirements. Uni-C can also be applied independently to enhance the output quality of individual generative models. Extensive experiments across video and 3D generation tasks demonstrate that Uni-C consistently improves generation quality in both workflow-based and standalone settings, highlighting its versatility and strong generalization capability.
\end{abstract}

\vspace{-0.5em}
\section{Introduction}
\label{sec:intro}

Workflow-based generation, which connects various generative models in a flow-like manner to accomplish diverse generation tasks, has garnered significant momentum within the AIGC community. Platforms such as ComfyUI~\cite{comfyui2023} and StableSwarmUI~\cite{stableswarmui2023} have achieved broad acceptance due to their user-friendly interfaces, which facilitate the customization of pipelines. These customizable pipelines empower users to create tailored workflows with ease, accommodating a variety of modalities and generation tasks~\cite{rombach2022high}.

However, the sequential use of different generative models often presents several challenges, including the amplification of cascading errors and domain discrepancies, which can result in suboptimal outcomes. This issue is especially pronounced in cross-modal applications, for example, when a video or 3D generative model follows an image generative model.

\begin{figure}[!t]
    \centering
    \includegraphics[width=0.5\textwidth]{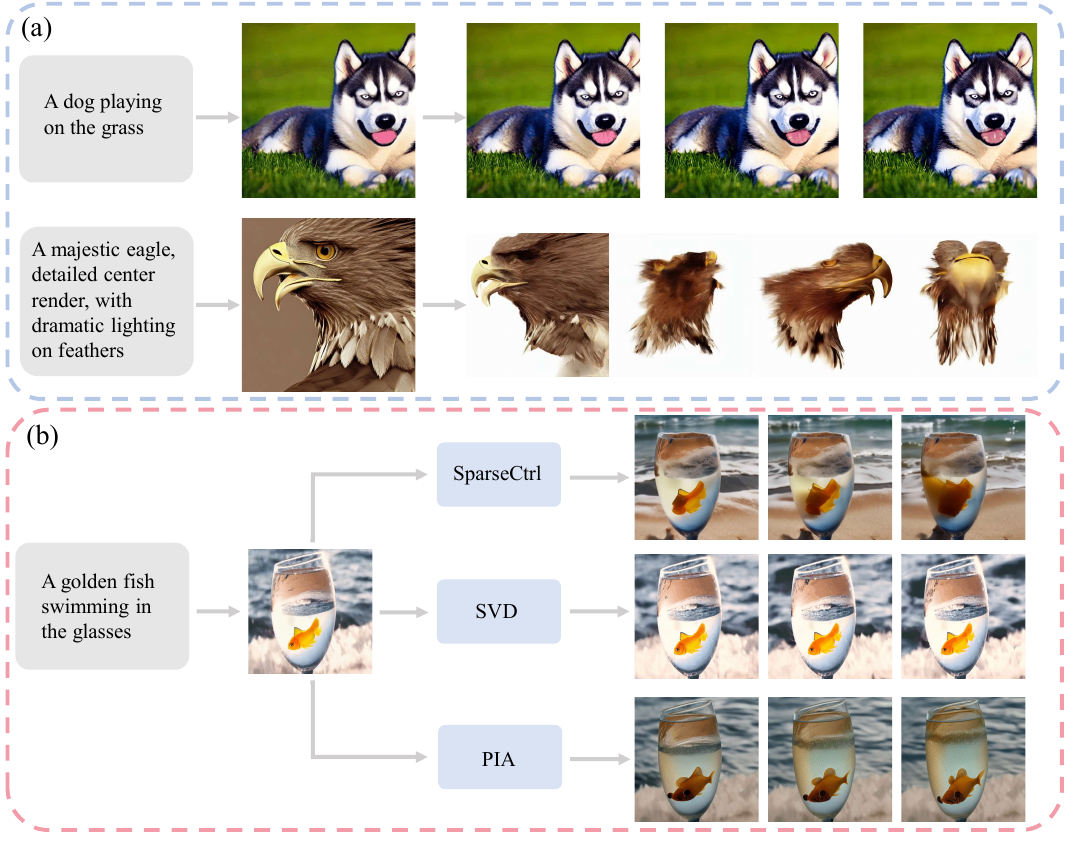}
    \caption{\textbf{Some generated images lead to substandard results for subsequent models.} (a) Explicit patterns. Positions or postures with a low likelihood of movement result in static scenes or only camera motion. Incomplete objects lead to distorted 3D objects. (b) Implicit patterns. Models generate nearly static or corrupted videos from visually plausible images.}
    \label{fig:issues}
    \vspace{-2em}
\end{figure}

There are two types of patterns in the generated images that can lead to substandard results for subsequent models. The first type consists of explicit patterns, as illustrated in Fig.~\ref{fig:issues}(a). Initial positions or postures with a low likelihood of movement often cause subsequent video generative models to produce outputs that are nearly static or exhibit only camera motion.  Besides, Incomplete foreground elements may lead to subsequent 3D generative models creating blurred or distorted 3D objects. The second type pertains to implicit patterns in the generated images that are not perceptible to the human eye. We have observed that video diffusion models often struggle to understand the semantics within these images, leading to difficulties in generating object motion. This issue is widespread across various video diffusion models in Fig.~\ref{fig:issues}(b).

To achieve satisfactory results in the workflow, users must invest significant effort in selecting suitable models and carefully cherry-pick the generated images for subsequent models. Alternatively, the entire workflow can be fine-tuned in an end-to-end manner. However, this approach is not only time-consuming but also challenging to adapt to a broad range of diverse generative tasks. To address these issues, we propose using the prior of the subsequent model to guide the generation process of the preceding model, ensuring the output distribution aligns with the input distribution of the subsequent model. 

To achieve this goal, we develop a universal classifier, referred to as Uni-Classifier (Uni-C), that incorporates generalizable priors for a variety of workflow-based generation tasks. Specifically: 1) We select the \textbf{video diffusion prior} as a generalizable prior due to the versatility inherent in the video modality. Images can be considered as videos with a single frame, and 3D objects, when depicted from multiple perspectives, can be viewed as a video sequence captured around the object. 2) We have established an \textbf{automatic and comprehensive pipeline} to generate pairs of videos and quality scores, thereby materializing the video diffusion prior. This pipeline involves feeding image samples generated by an image diffusion model into a video diffusion model and then using LVLM (Large Vision Language Model) to assess the quality of the generated videos. 3) We train our universal classifier using the paired data generated from this pipeline.

Though data generation workflow includes an image diffusion model (SD~\cite{rombach2022high} model) and a video diffusion model (SVD~\cite{blattmann2023stable} model), Uni-C exhibits a high degree of generalizability and can be seamlessly integrated into a multitude of workflows in a plug-and-play manner.
Beyond workflow-level generation, Uni-C can also be effectively utilized in individual model generation across various video, and 3D diffusion models, 
thereby enhancing the quality of generation.
We attribute this effectiveness to the versatility inherent in the video diffusion prior 
that has been incorporated into the classifier. The contributions are summarized as follows:

\begin{itemize}






\item We introduce the Uni-C, leveraging video diffusion priors to guide the denoising process, thereby effectively narrowing the gap in workflow-based generation.


\item Uni-C is a versatile, plug-and-play module with extensive applicability. It excels in both workflow-level and individual model generation, supporting a variety of generation models for images, videos, and 3D content.

\item Extensive experiments demonstrate our Uni-C achieves superior quantitative and qualitative results on various benchmarks, serving as an effective bridging model for generative models and workflows tasks.

\end{itemize}
\section{Related Work}
\label{sec:related}

\noindent\textbf{Image and Video Synthesis with Diffusion Models.}
Stable Diffusion (SD) and its variants have become the backbone of text-to-image synthesis, enabling diverse applications such as subject-driven generation and instruction-based manipulation~\cite{rombach2022high, zhang2023adding}. However, purely text-guided image generation often yields semantically incomplete or poorly integrated objects, especially when downstream tasks require strong motion or structural cues that are absent in static images.
Video generation methods typically extend SD via temporal adapters, falling into image-to-video~\cite{blattmann2023stable, zhang2024pia, guo2025sparsectrl} and text-to-video~\cite{singer2022makeavideo, guo2023animatediff, chen2024videocrafter2} paradigms. The former relies heavily on the motion readiness of input frames, while the latter suffers from temporal jitter and object inconsistency due to a lack of inter-frame guidance. To address these limitations across both settings, our Uni-C leverages video diffusion priors to guide SD-based image generation, producing motion-aware frames for image-to-video pipelines and enhancing temporal coherence when integrated into T2V models like AnimateDiff~\cite{guo2023animatediff}.

\noindent\textbf{Workflow-Based Generation.}
Workflow-based generation has gained traction in the AIGC community through platforms like ComfyUI~\cite{comfyui2023} and StableSwarmUI~\cite{stableswarmui2023}, which enable users to compose modular pipelines by chaining blocks of diverse generative models. For example, connecting image synthesis to video synthesis modules. However, such workflows often suffer from distributional misalignment between upstream outputs and downstream inputs, requiring manual tuning of models or parameters to ensure compatibility. To address this, our Uni-C leverages video diffusion priors to construct a classifier that guides the upstream diffusion process toward outputs better aligned with downstream video generation requirements, thereby reducing the need for ad-hoc adjustments and improving end-to-end quality.
\section{Method}
\label{sec:method}

\begin{figure*}[!ht]
    \centering
    \includegraphics[width=0.85\textwidth]{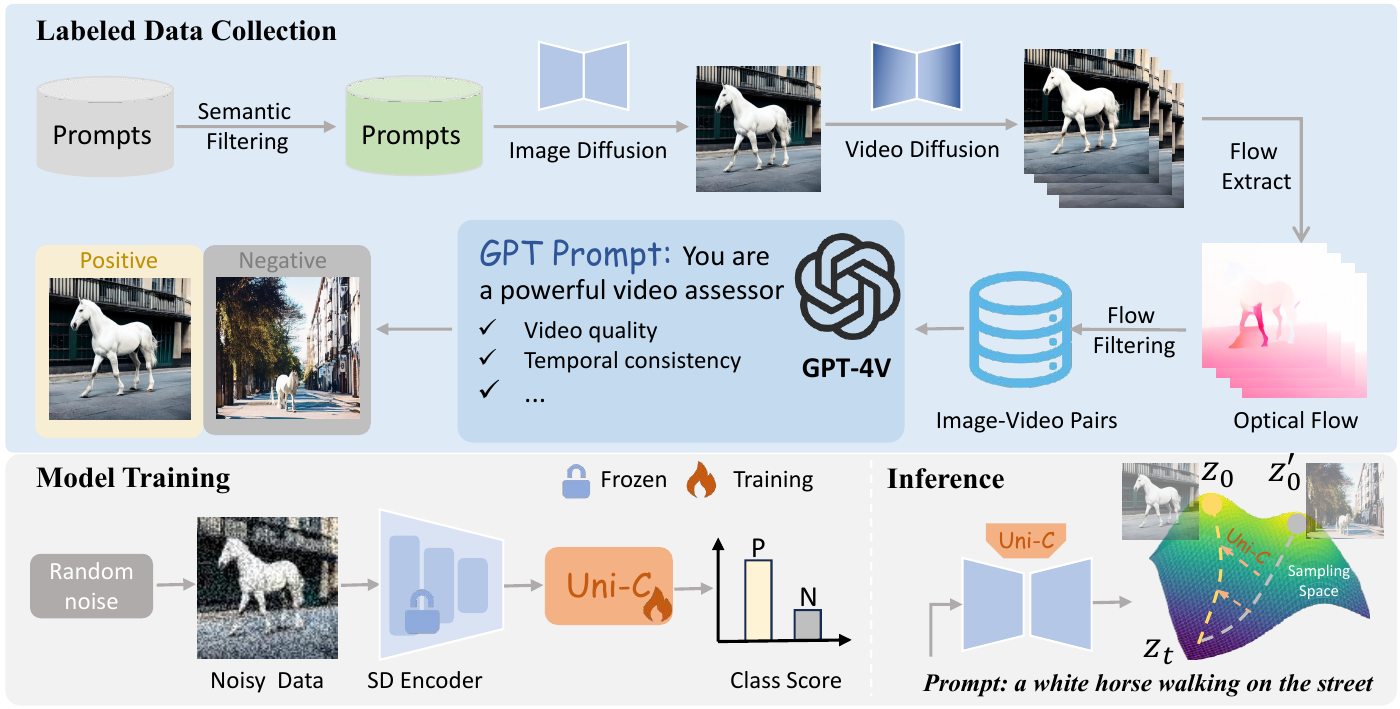}
    \caption{\textbf{The pipeline of Uni-C.} The upper part illustrates the data collection process and the use of LVLM knowledge to create positive and negative samples. The lower part shows the training and inference process of the Uni-C. }
    \label{fig:pipe}
    \vspace{-1em}
\end{figure*}

\subsection{Preliminaries}
\label{sec:pre}

{\noindent\bf Stable Diffusion.}
Our Uni-C is built on a pretrained Stable Diffusion (SD)~\cite{rombach2022high} model, 
where an autoencoder $\{\mathcal{E}(\cdot), \mathcal{D}(\cdot)\}$ is applied to 
encode input image $x$ into latent space $z=\mathcal{E}(x)$, 
and a conditional diffusion model $\epsilon_\theta(\cdot)$ is utilized to 
map the noised image latent $z_t$ to pure latent $z_0$, where $t$ is the time step 
varies form $0$ to $T$.
\cite{ho2020denoising} note that $z_t$ can be expressed as a Gaussian distribution:

\begin{equation}
\begin{aligned}
z_t & =\mathcal{N}\left(z_t ; \sqrt{\bar{\alpha}_t} z_0,\left(1-\bar{\alpha}_t\right) \mathbf{I}\right) \\
& =\sqrt{\bar{\alpha}_t} z_0+\epsilon \sqrt{1-\bar{\alpha}_t}, \epsilon \sim \mathcal{N}(0, \mathbf{I}).
\end{aligned}
\end{equation}
Here, $1-\bar{\alpha}_t$ is the variance schedule that tells the variance of the noise.
To restore $z_0$ from $z_t$, the diffusion model is trained to estimate the noise component, which can be formulated by the following objective function:

\begin{equation}
\mathcal{L}(\theta)=\mathbb{E}_{ \epsilon , t }\left[\left\|\epsilon-\epsilon_\theta\left(z_t, c, t\right)\right\|_2^2\right],
\end{equation}
where $c$ is the condition prompt. 

{\noindent\bf Classifier guidance.}
A notable feature of diffusion models is the ability to incorporate 
constraints during inference to guide the sampling direction $\hat{\epsilon}_t$ at each iteration.
Classifier guidance~\cite{dhariwal2021diffusion} is one of such methods, 
introduced to lead the image generation to a specific class $y$.
To achieve this, we need to construct a labeled dataset and 
pretrain a classifier $p(y | z_t)$ on the noisy image latent $z_t$.
During sampling, the gradient
of the classifier is used to guide $\hat{\epsilon}_t$:

\begin{equation} \label{eq:guide} 
\hat{\epsilon}_t=\epsilon_\theta\left(z_t, c, t\right)- \lambda  \sqrt{1-\bar{\alpha}_t}  \nabla_{z_t} \log p\left(y \mid z_t\right),
\end{equation}
where $\lambda$ is the guidance weight.

\subsection{Observation and Analysis}

To handle complex and diverse generation tasks, workflow-based generation, which sequentially integrates various generative models, has gained significant traction within the AIGC community. However, certain patterns in the outputs of preceding models can result in substandard performance for subsequent models, as shown in Fig.~\ref{fig:issues}. To identify the commonalities of these issues and seek universal solutions, we conducted the following analysis.

\begin{figure}[!ht]
    \centering
    \includegraphics[width=0.5\textwidth]{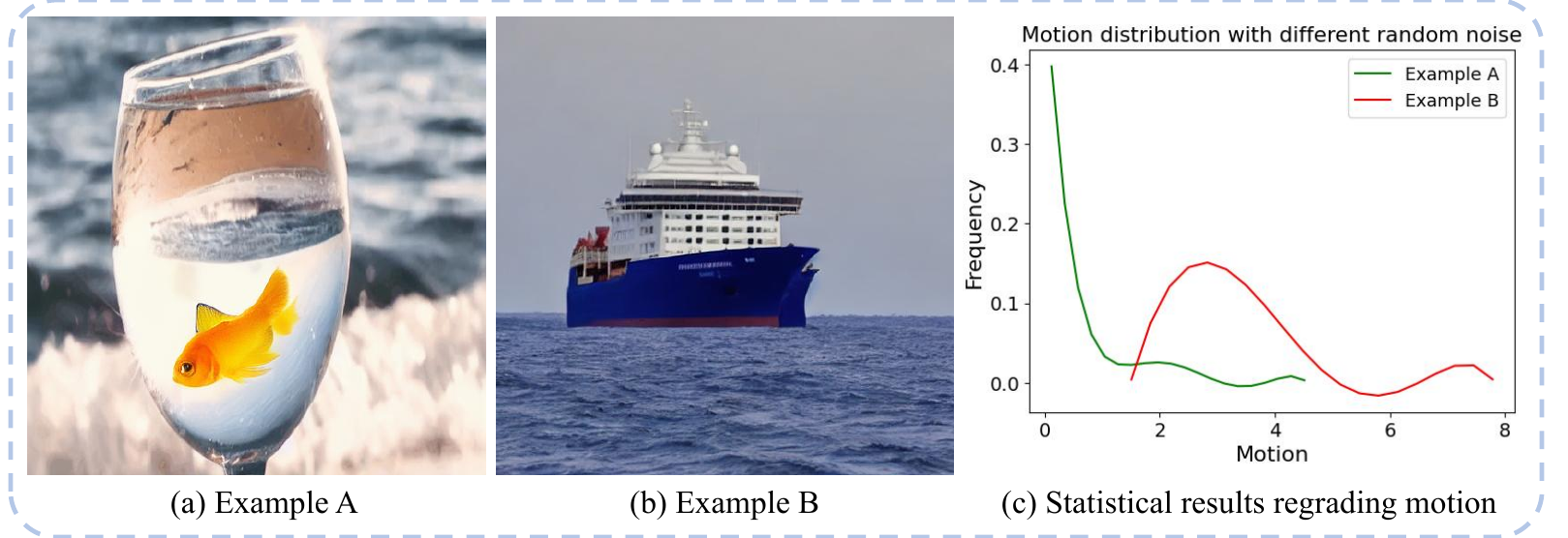}
    \vspace{-1em}
    \caption{Statistical results regarding motion in the generated videos (I2V) with 100 random initial noises, in which the spatial mean of the motion has been subtracted.}
    \vspace{-1.5em}
    \label{fig:obs}
\end{figure}

\noindent\textbf{Video modality.} The video modality is relatively universal. Images can be seen as single-frame videos, and 3D objects, when viewed from multiple angles, can be treated as a video sequence captured around the object. A universal solution should leverage the priors of video diffusion models.

\noindent\textbf{Video diffusion models.} When using generated images featuring patterns as depicted in 
Fig.~\ref{fig:issues} as inputs, the videos generated from various video diffusion models with notable similarities:
all three models fail to produce realistic motion for the goldfish in the glass, 
leading to either static objects or distorted movement.
Thus, it is possible to generalize the priors from one video diffusion model to others.

\noindent\textbf{Sampling noise.} When given input images and a video diffusion model, different sampling noises during the inference phase generate a similar motion. 
To prove it, we analyze motion distribution to reveal the pattern of image motion variation across 100 random cases.
As shown in Fig.~\ref{fig:obs}, example B exhibits a greater variation relative to the average optical flow, 
making it statistically more suitable as a reference image for the image-to-video model

Based on the analysis, leveraging the prior knowledge from a single video diffusion model to construct a universal classifier presents a potentially feasible strategy 
for addressing issues in workflow-based generation.


\subsection{Labeled Dataset Collection}
\label{sec:data}
As a special classifier guidance method, our Uni-C is designed to guide the diffusion process in directions that are more favorable for downstream video generation. 
Here, the class label $y$ 
is not confined to a specific category.
Instead, it refers broadly to images that serve as effective inputs for video generation, 
by contributing to outputs that exhibit realistic and coherent object motion.
To this end, we first design a data generation pipeline (Fig.~\ref{fig:pipe}) to create an image-video pair dataset and then label each pair with LVLM knowledge.

{\noindent\bf Image-Video pair generation.}
As shown in the upper half of Fig.~\ref{fig:pipe}, to collect a comprehensive image-video pair dataset from real-world scenarios, we utilized VidProM~\cite{wang2024vidprom}, a large-scale dataset containing text prompts from real users. A semantic de-duplication algorithm is applied to ensure that the cosine similarity between any two arbitrary prompts remains below $0.8$.
Subsequently, we use the SD model to generate five distinct images per prompt,
which serve as initial frames and are then input into a video diffusion model 
to generate corresponding videos.

{\noindent\bf Labeling with LVLM knowledge.}
In the labeling phase, a two-stage approach is applied
to evaluate the motion quality of each video, 
which guides the labeling of the corresponding images. 
First, we calculate the mean and variance of the optical flow for each video, 
assigning a negative label to those with minimal optical flow variation.
Inspired by ~\cite{tao2024motioncom,yang2024mastering},
A multi-model LVLM, specifically GPT-4V, is then utilized to strategically assess the videos based on three criteria: 
object completeness, temporal consistency, and clear motion direction for the object. 
A positive label is assigned to the corresponding image only if all requirements are met; otherwise, a negative label is applied.
With the assistance of LVLM knowledge, the motion cues of positively labeled videos are further ensured.

\subsection{Uni-Classifier}
\label{sec:uni-c}

\label{sec:experiment}
\begin{figure*}[!ht]
    \centering
    \includegraphics[width=0.95\textwidth]{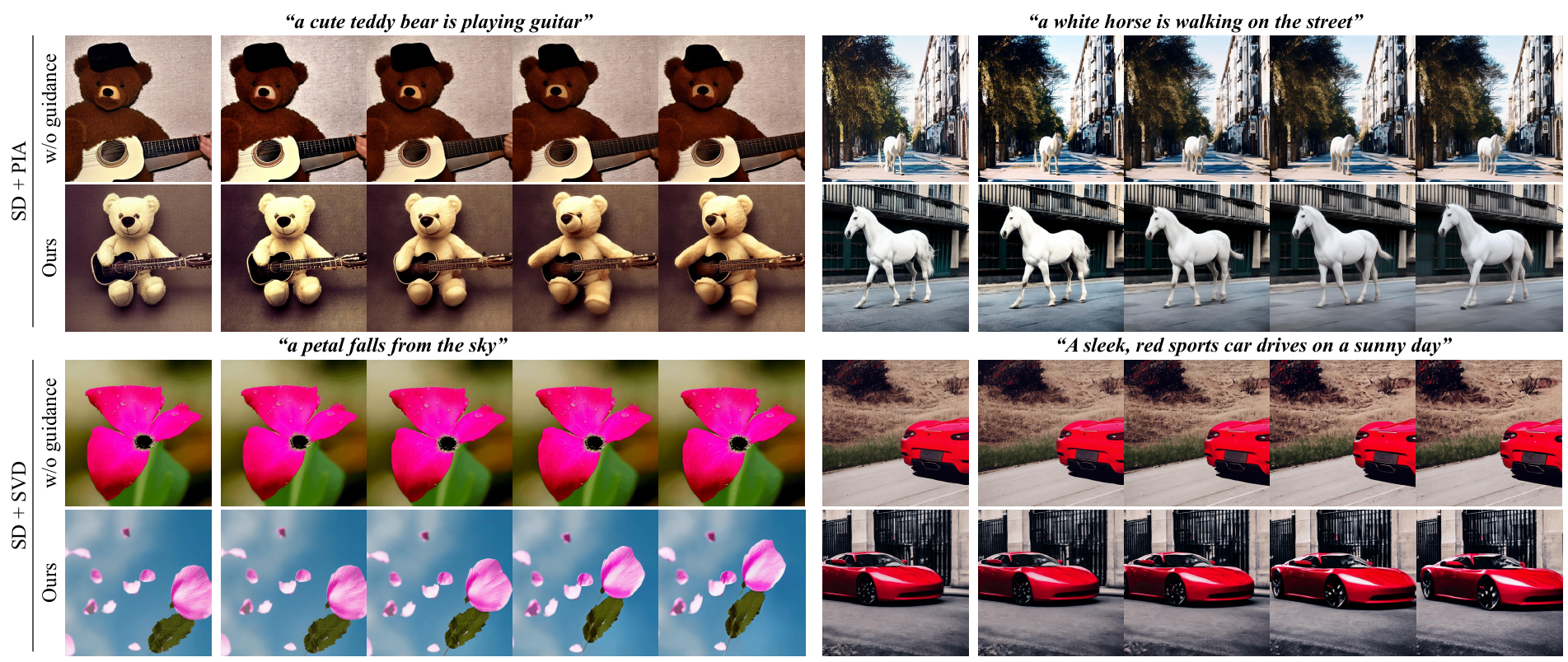}
    \vspace{-0.5em}
    \caption{\textbf{Visual comparison in workflow-based video generation.}}
    \label{fig:i2v}
    \vspace{-1.5em}
\end{figure*}

\begin{figure*}[!ht]
    \centering
    \includegraphics[width=0.95\textwidth]{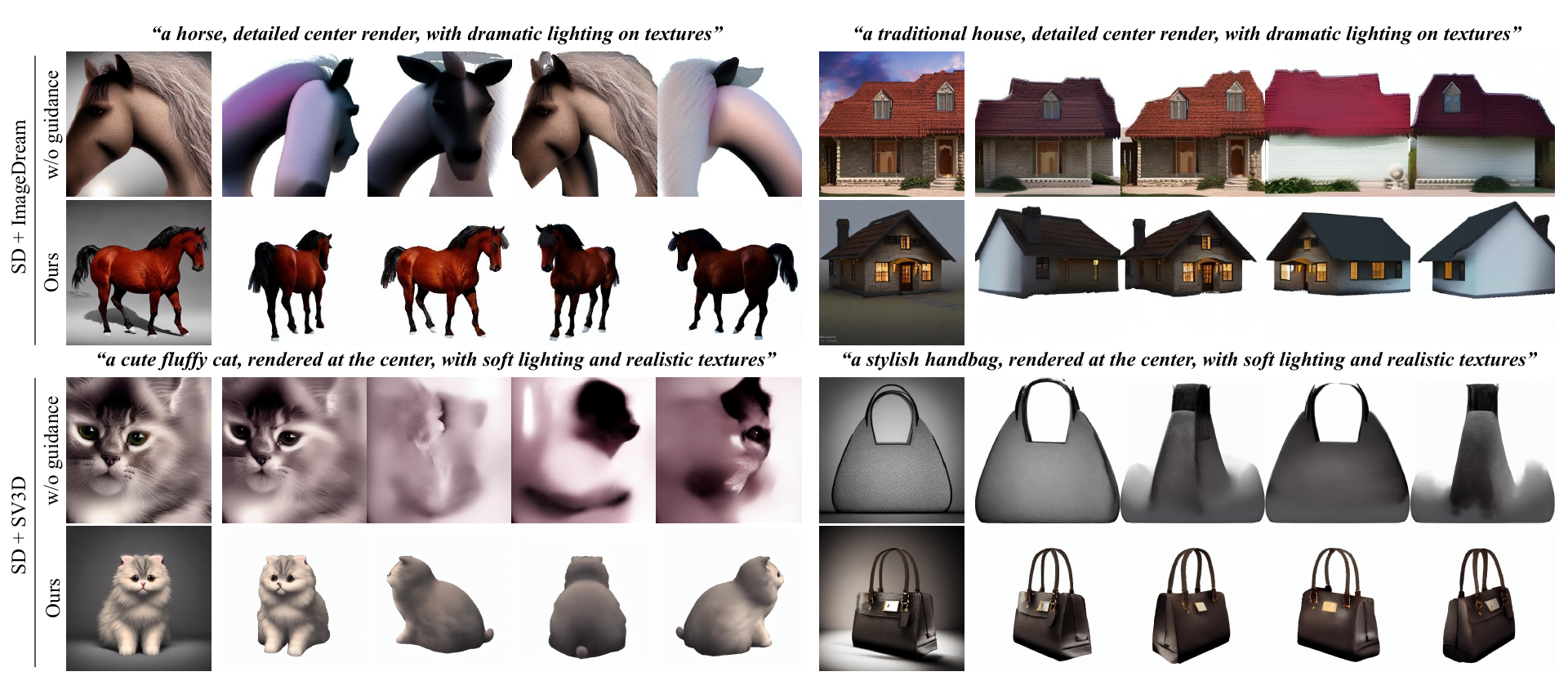}
    \vspace{-0.5em}
    \caption{\textbf{Visual comparison in workflow-based 3D generation.}}
    \vspace{-1.5em}
    \label{fig:i23d}
\end{figure*}

{\noindent\bf Model structure.}
With the labeled dataset, our Uni-C aims to incorporate video priors into the diffusion generation process via classifier guidance.
As shown in the lower half of Fig.~\ref{fig:pipe}, a U-Net Encoder in SD is utilized to extract the features of the input noisy latent $z_t$.
To distinguish the high-level semantics of the latent, 
a simple binary classifier with an average pooling layer is employed to produce the final logits scores.
For the training phase, we keep the U-Net Encoder fixed to retain its representation capacity,
and entirely train the classifier.
During the denoising process, the scaled gradient 
$\lambda \nabla_{z_t} \log p\left(y \mid z_t\right)$
of the classifier is used to guide the sampling process from the negative image $z_0^{'}$
towards a positive labeled image $z_0$.

{\noindent\bf Downstream tasks.}
As a plug-and-play module, our once-trained Uni-C serves as a bridge between image generation and downstream tasks. 
It provides reference images with enriched motion cues for image-to-video and image-to-3D models. 
Additionally, Uni-C can be directly embedded into text-to-video and text-to-3D tasks. 

\section{Experiments}
\begin{figure*}[!ht]
    \centering
    \includegraphics[width=0.95\textwidth]{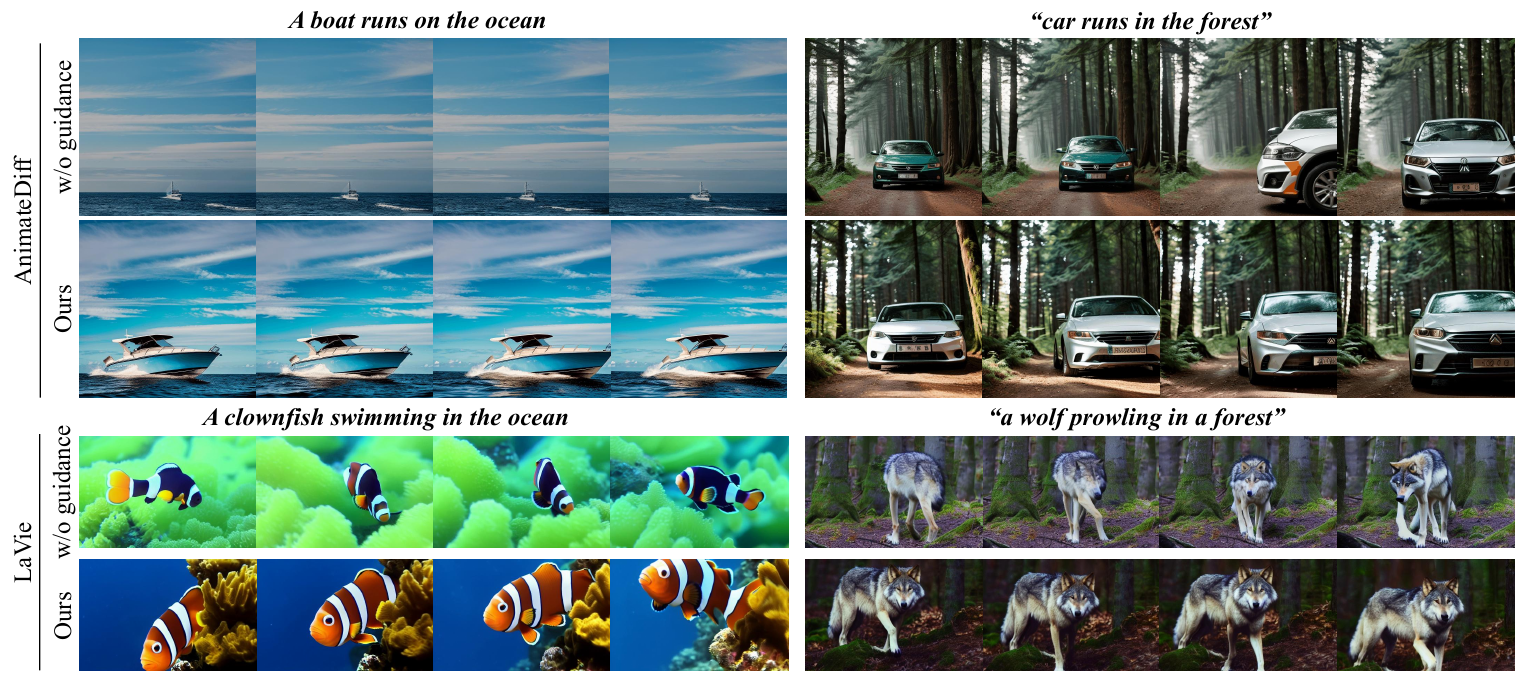}
    \vspace{-0.5em}
    \caption{\textbf{Visual comparison in individual text-to-video generation.}}
    \vspace{-1.5em}
    \label{fig:t2v}
\end{figure*}

\begin{figure}[!t]
    \centering
    \vspace{-0.9em}
    \includegraphics[width=0.45\textwidth]{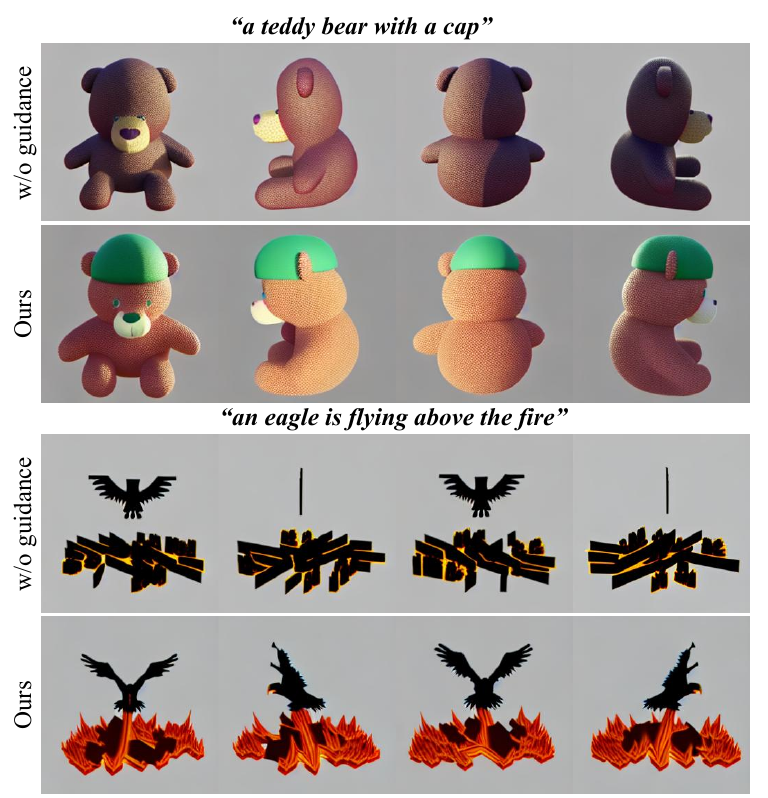}
    \caption{\textbf{Visual comparison in text-to-3D generation.}}
    \vspace{-2em}
    \label{fig:t23d}
\end{figure}

\subsection{Implementation Details}
In this work, we use the VidProM~\cite{wang2024vidprom} dataset,
which provides 8,406 user-preference text prompts after prompt filtering. 
Using the SD~\cite{rombach2022high} text-to-image model, 
each prompt generates five different images. 
These images serve as reference frames and are then input into the
Stable Video Diffusion~\cite{blattmann2023stable} image-to-video model to obtain corresponding generated videos.
In total, we collect about 40k image-video pairs, and then GPT-4V~\cite{achiam2023gpt} is applied to label the dataset.
The classifier is trained on a single A100 GPU for 20,000 steps with a learning rate of $1e-5$.
Once trained, the same classifier can be embedded directly into the sampling processes of various downstream tasks.
For each task, DDIM sampler~\cite{song2020denoising} with Uni-C is adopted in the inference phase.

\vspace{-0.3em}
\subsection{Experimental Setup}
{\noindent\bf Video generation methods.}
For video generation tasks, we evaluate the performance of Uni-C using VBench~\cite{huang2023vbench}, 
with Dynamic Degree, Aesthetic Quality, and Imaging Quality as evaluation metrics.
Specifically, for the workflow-based video generation method, 
the prompts are input into the SD model, 
and the generated images are used as reference images for the subsequent video generation model,
such as Stable Video Diffusion~\cite{blattmann2023stable} and PIA~\cite{zhang2024pia}.
For the individual text-to-video model, such as Animatediff~\cite{guo2023animatediff} and LaVie~\cite{wang2023lavie}
the prompt is directly used as the conditioning input for text-to-video generation.
Note that "Image Priors" refers to the approach of bypassing the image-to-video model 
and directly using GPT-4V to label the generated images.
This setup serves as an ablation study, demonstrating the importance of video prior knowledge for workflow tasks.

{\noindent\bf 3D generation methods.}
For the 3D generation tasks,
the prompts are from GPTEval3D~\cite{wu2024gpt} Benchmark.
As an evaluation metric for text-to-3D generative models, 
GPTEval3D provides $110$ diverse prompts.
Similarly, for the workflow-based 3D generation method, the SD model takes an input prompt and 
then feeds the generated image into a downstream image-to-3D model,
such as SV3D~\cite{voleti2025sv3d} and ImageDream~\cite{wang2023imagedream}.
The individual text-to-3D model MVDream~\cite{shi2023mvdream} takes a text prompt as input.
For objective evaluation, we follow Q-ALIGN~\cite{wu2023q} and use the 
i) Image Quality Assessment (IQA) and ii) Image Aesthetic Assessment (IAA) to evaluate 
the average image quality and average aesthetic score of multi-view images, respectively.
The Clip Score is also used to calculate the average text alignment score.


\vspace{-0.5em}
\subsection{Qualitative Comparison}


\begin{table}[!t]
\centering
\footnotesize
\caption{VBench metrics for video generation.}
\resizebox{0.98\linewidth}{!}{
\begin{tabular}{lccc}
\toprule
Method                & Dynamic Degree & Aesthetic Quality  & Imaging Quality\\ \midrule
  SD$\rightarrow$SVD          & 0.0250 & 0.5978 & 0.6859     \\
    + Image Priors    & 0.0320 & 0.5998 & 0.6861              \\
 \rowcolor{gray!20} \textbf{+ Uni-C}             &\textbf{0.0500} & \textbf{0.6113} & \textbf{0.6921}     \\ \midrule
 SD$\rightarrow$PIA               & 0.0500 & 0.6344 & 0.7110     \\
     + Image Priors   &  0.0675      &  0.6355        &  0.7177               \\
 \rowcolor{gray!20}\textbf{+ Uni-C}              &\textbf{0.1250} &\textbf{0.6408} & \textbf{0.7191}    \\ \midrule
 AnimateDiff          & 0.0625 & 0.6500 & 0.7337     \\
     + Image Priors    & 0.0750      & 0.6508         &   0.7335              \\
 \rowcolor{gray!20} \textbf{+ Uni-C}            &  \textbf{0.1500} & \textbf{0.6673} & \textbf{0.7471}   \\ \midrule
 LaVie                & 0.4375 & 0.5705 &0.6843     \\
     + Image Priors    & 0.3875      &  0.5746        &  0.6846                 \\
 \rowcolor{gray!20} \textbf{+ Uni-C}             &  \textbf{0.6000}    &\textbf{0.5896} &\textbf{0.6992}     \\ \bottomrule
\end{tabular}
}
\vspace{-1.5em}
\label{tab:vbench}
\end{table}

\begin{table}[!t]
\centering
\footnotesize
\caption{Quantitative metrics in 3D generation quality.}
\begin{tabular}{lccc}
\toprule
 Method               & IQA $(\uparrow)$    & IAA $(\uparrow)$ & Clip Score $(\uparrow)$ \\ \midrule
SD$\rightarrow$SV3D & 3.2246 & 2.9141  & 28.15    \\
\rowcolor{gray!20} \textbf{+ Uni-C}              & \textbf{3.3272} & \textbf{3.0156}   & \textbf{28.93}   \\ 
SD$\rightarrow$ ImageDream               & 2.8164 & 2.2656  & 28.10    \\
\rowcolor{gray!20} \textbf{+ Uni-C}              & \textbf{2.9102} & \textbf{2.3555} & \textbf{28.22}    \\ \midrule  
 MVDream                & 2.4668 & 1.6250  & 26.57   \\
\rowcolor{gray!20}\textbf{+ Uni-C}              &  \textbf{2.5883} &  \textbf{1.7260} &  \textbf{27.82}     \\ \bottomrule
\end{tabular}
\vspace{-2em}
\label{tab:3d_metric}
\end{table}

{\noindent\bf Workflow-based tasks.}
Fig.~\ref{fig:i2v} shows the workflow-based video generation results. 
While the vanilla SD model aligns with the prompt’s visual semantics, 
issues with positioning, postures, and occasional incomplete objects hinder its integration with downstream image-to-video models.
With the guidance of our Uni-C, 
the motion cues of objects in the generated images are enhanced, 
improving both the range and quality of object movement in the resulting video.
For workflow-based 3D generation results as shown in the Fig.~\ref{fig:i23d}, 
the completeness of objects in the generated image is crucial.
Without guidance, despite the prompt specifying position and lighting conditions, 
the SD model still struggles to generate suitable reference images for the image-to-3D model.


{\noindent\bf Individual tasks.}
Our Uni-C can also be used in individual model generation in a plug-and-play manner.
As shown in Fig.~\ref{fig:t2v}, the vanilla text-to-video model 
suffers from object collapse or replacement during motion generation. 
Our Uni-C mitigates this issue by providing frame-by-frame guidance, 
enhancing object completeness and temporal consistency.
Similarly, for the text-to-3D tasks as shown in Fig.~\ref{fig:t23d},
multi-view generation suffers from object collapse in certain perspectives, 
leading to planar results.
Uni-C mitigates it by providing guidance for each perspective, 
promoting more complete object generation in line with the text input conditions.

\subsection{Quantitative Comparison}
We present quantitative results for video and 3D generation in Tab.~\ref{tab:vbench} and Tab.~\ref{tab:3d_metric}, respectively. 
Under the guidance of our Uni-C, both tasks achieve consistently higher quality and aesthetic scores. 
Moreover, the generated outputs exhibit stronger alignment with the input text prompts.
Notably, the enhancement in the Dynamic Degree is particularly significant,
demonstrating its effectiveness in improving motion quality.
\section{Conclusion}
\label{sec:conclusion}
In this work, we propose Uni-C, which utilizes video diffusion priors to address the domain gap 
between upstream 2D AIGC tasks and downstream video or 3D generation. 
As a plug-and-play module, Uni-C can be applied 
to a variety of downstream tasks, 
including text-to-video and text-to-3D.
We offers the AIGC community an available tool for 
effectively connecting different blocks in the 
workflow generation.

\bibliographystyle{IEEEbib}
\bibliography{icme2026references}

\clearpage
\setcounter{page}{1}
\maketitlesupplementary
\section{Detailed Data Collection Processes}
In this section, we provide a detailed explanation of how labeled image and video data pairs
are constructed using Video Diffusion Priors with the assistance of GPT-4V~\cite{achiam2023gpt}.
{\noindent\bf Prompts collection.}
VidProM~\cite{wang2024vidprom} contains 1.67 million unique text-to-video user-preference prompts. 
Firstly, a semantic de-duplication algorithm is applied to ensure 
that the cosine similarity between any two prompts remains below 0.8. 
Next, a Qwen-72b~\cite{bai2023qwen} LLM model is used to filter the prompts further, 
removing non-English, unclear prompts, 
and ensuring the inclusion of motion-related words to avoid purely static prompts.
In the end, we collected 8,406 unique prompts that meet all criteria.

{\noindent\bf Image-Video pairs collection.}
After collecting the prompts, we use the classical Stable Diffusion (SD)~\cite{rombach2022high} model,
the most widely used SD model with support for various community-driven extensions,
to generate five images for each prompt with random seeds. 
Then all images serve as the reference images and are input into Stable Video Diffusion(SVD)~\cite{blattmann2023stable} model to generate corresponding videos.
As a result, we obtained about 40k image-video pairs for the subsequent labeling process.

{\noindent\bf LVLM labels collection.}
First, we use a flow-based filtering method for the initial labeling process of the videos. 
For each video, we extract the mean and variance of the horizontal, vertical, and overall optical flow. 
Videos are labeled as negative if the overall optical flow mean is less than its variance, 
or if both the horizontal and vertical optical flow means are less than five times their respective variances.
Otherwise, the remaining videos proceed to the next phase, where GPT-4V is used for further filtering.

With the help of LVLM, we defined a set of criteria to evaluate video quality, 
temporal consistency, and motion amplitude. 
Only videos that meet all the conditions are assigned a positive label.
We extract four frames from each video at regular intervals, starting with the first frame,
and then use GPT-4V for labeling. Below are the prompts provided to GPT-4V for evaluation:

\begin{itemize}
\item The objects are visible, and their size takes up at least 10\% of the image.
\item None of the objects should appear distorted, and their appearance must be consistent with real-world scenarios.
\item The features of the object(s) should remain reasonably consistent in all frames.
\item The movement or motion of the object should be detected. 
\item Motion limited to camera panning or zooming is unacceptable.
\end{itemize}

Specifically, points 1 and 2 constrain the prompts on the shape, completeness, and proportion of objects and ensures the overall quality of the objects.
Then, points 3 and 4 focus on temporal consistency across frames, 
requiring not only object motion in the video but also preventing object collapse.
Finally, pure camera motion is undesirable, so we exclude it.

Only videos that pass all filtering processes conducted by GPT-4V are assigned a positive label, ultimately guiding the gradient direction during the SD model's inference phase. 
Notably, during the labeling process, we randomly sampled 1,000 cases initially 
labeled as positive by GPT-4V for manual verification by experts.

\section{The Studies of Parameter $\lambda$}
\label{sec:paras}

\subsection{Parameter Settings}
Tab.~\ref{tab:params} shows the selection of parameter $\lambda$ for quantitative experiments, 
including video generation and 3D generation tasks.
A larger value indicates stronger guidance of the Uni-C on the model.
For workflow-based video generation tasks, the upstream SD model's ability to generate objects with richer motion cues is crucial 
for enabling downstream models like SVD and PIA to produce high-quality videos. 
Therefore, a larger $\lambda$ is applied.
In contrast, for individual text-to-video generation tasks, 
such as Animatediff and LaVie, where models are inherently trained on video data, their capacity to generate dynamic videos is generally stronger,
allowing for the use of a smaller $\lambda$.
For 3D tasks, due to the domain gap between our Uni-C training data and 3D input data, 
a larger $\lambda$ is utilized to ensure the generation of complete objects suitable for multi-view reconstruction.

\begin{table}[!t]
\centering
\caption{Guidance weight $\lambda$ selection in quantitative experiments.}
\begin{tabular}{lc}
\toprule
Task                        & $\lambda$ \\\midrule
SD$\rightarrow$SVD~\cite{blattmann2023stable}       & 8      \\
SD$\rightarrow$PIA~\cite{zhang2024pia}        & 8      \\
Animatediff~\cite{guo2023animatediff}                 & 2      \\
LaVie~\cite{wang2023lavie}                       & 1      \\\midrule
SD$\rightarrow$SV3D~\cite{voleti2025sv3d}       & 6      \\
SD$\rightarrow$ImageDream~\cite{wang2023imagedream} & 6      \\
MVDream~\cite{shi2023mvdream}                     & 8      \\\bottomrule
\end{tabular}
\label{tab:params}
\end{table}

\begin{figure}[!t]
    \centering
    \includegraphics[width=0.5\textwidth]{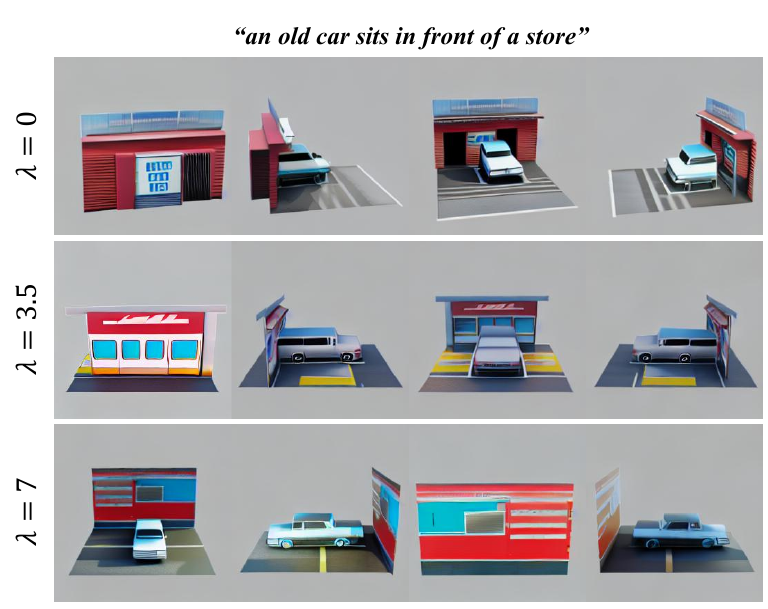}
    \vspace{-1.5em}
    \caption{Influence of different $\lambda$ on text-to-3D generation.}
    \vspace{-1em}
    \label{fig:analysis}
\end{figure}

\subsection{Ablation Studies}
In this section, we analyze a hard case with different guidance weights $\lambda$ on text-to-3D task.
As shown in Fig.~\ref{fig:analysis}, the text prompt describes both the foreground and background and their relative spatial relationship.
When $\lambda=0$, the generated car is incomplete and merged with the wall.
As $\lambda$ increases from $3.5$ to $7$, the car progressively recovers its complete form and ultimately detaches from the wall.
This transformation demonstrates the effectiveness of our Uni-C on object completeness and visual quality.

\section{Comparison with Image Diffusion Priors}
\label{sec:imgpriors}

\begin{figure}[!t]
    \centering
    \includegraphics[width=0.48\textwidth]{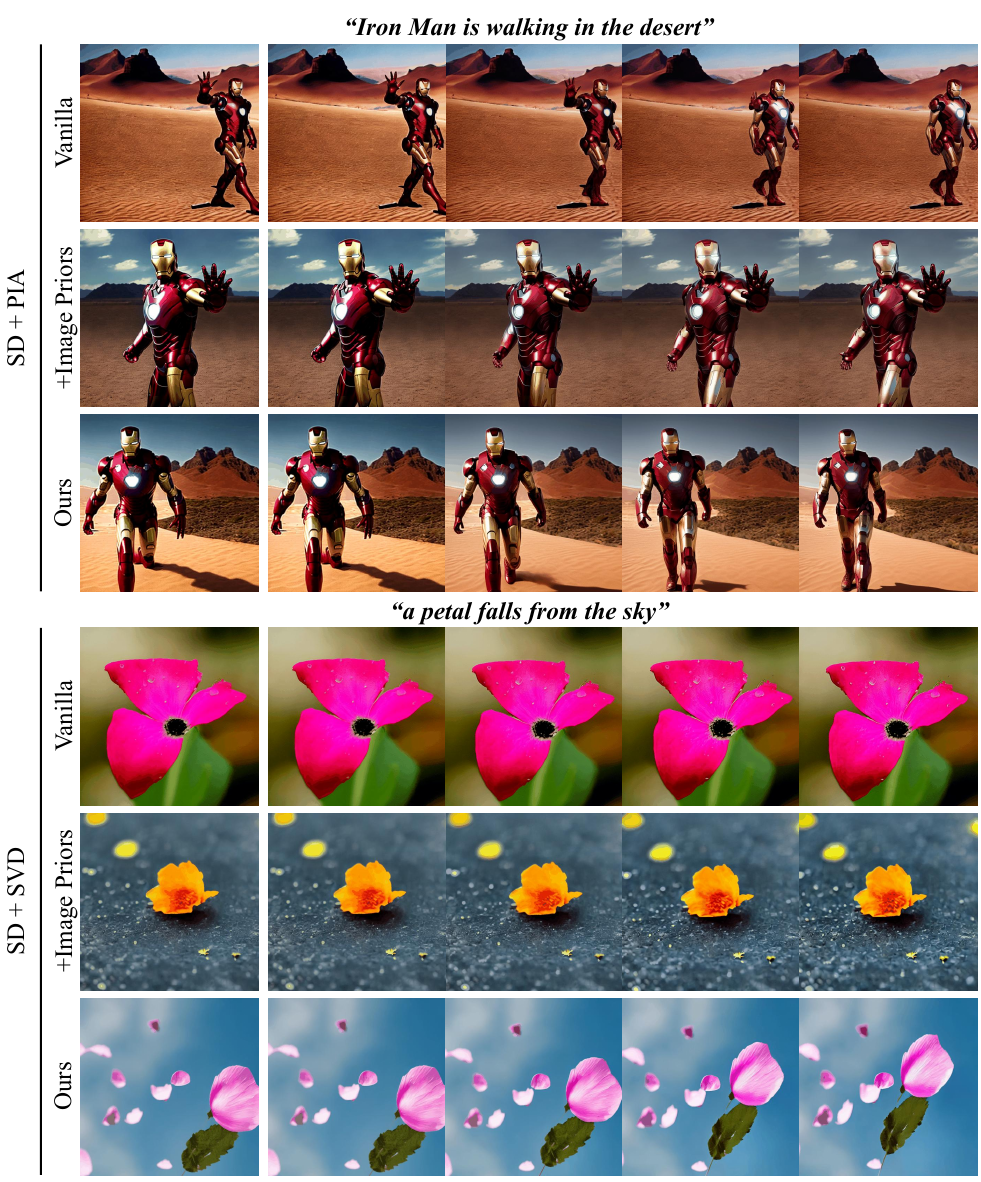}
    \caption{\textbf{Visual comparison with Image Priors model.} With the assistance of Image Priors, although the object's position is improved, 
    the overall posture remains unsuitable for subsequent motion generation.}
    \vspace{-1.3em}
    \label{fig:img_prior}
\end{figure}

\begin{figure*}[!ht]
    \centering
    \includegraphics[width=0.98\textwidth]{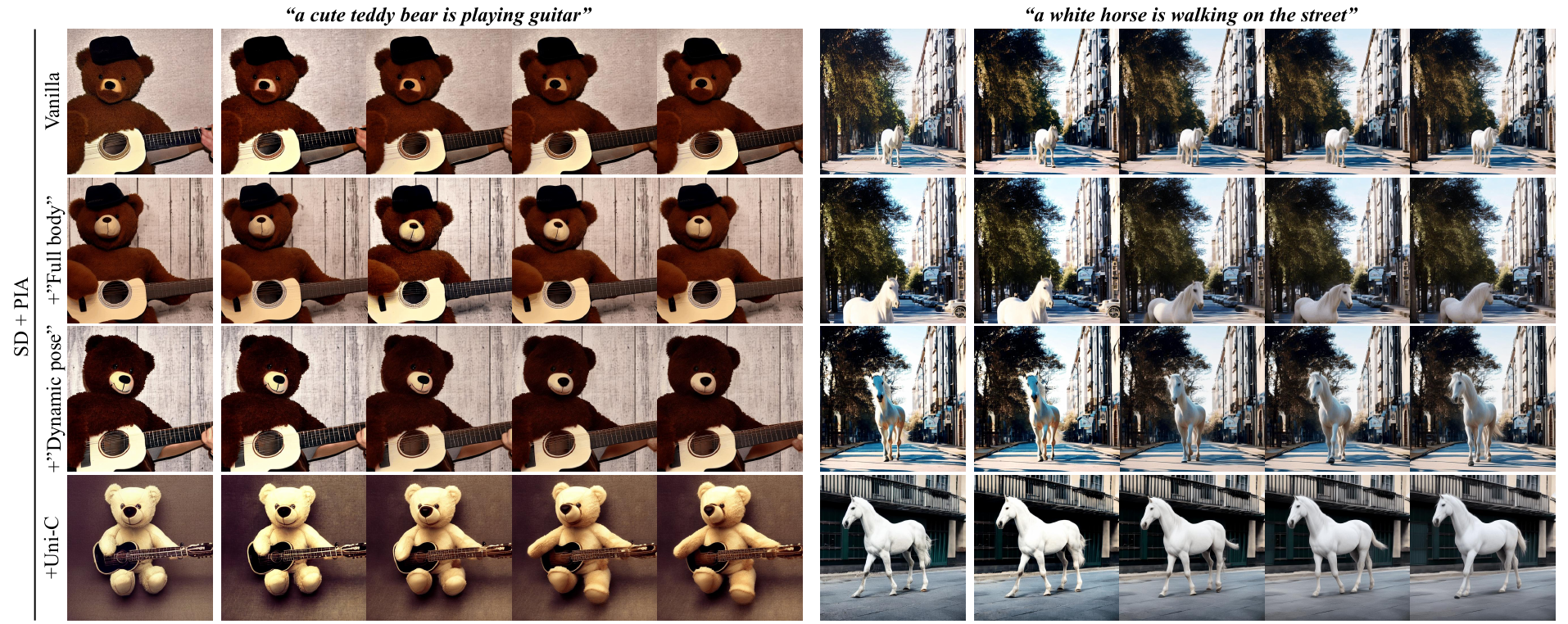}
    \vspace{-0.5em}
    \caption{\textbf{Visual comparison with prompt engineering.} Adding terms like "full body" and "dynamic pose" to the text prompt does not yield satisfactory results.}
    \vspace{-1em}
    \label{fig:prompt}
\end{figure*}
In this section, we analyze and compare the performance of using
Image Diffusion Priors versus Video Diffusion Priors. 
Specifically, during the labeling stage, images generated by SD are directly
evaluated by GPT-4V without undergoing the image-to-video process.
Then, positive labels are assigned based solely on image quality,
without considering the motion quality of the generated videos.
As shown in the Tab.I in the main paper, relying solely on Image Priors leads to limited improvement
in the Dynamic Degree metric for video generation, 
as it lacks constraints on both temporal consistency and motion intensity.

Similarly, we qualitatively demonstrate the results under the same prompt conditions
for the Vanilla SD model, the Image Priors model, and our Uni-C. 
As shown in Fig.~\ref{fig:img_prior}, the use of image priors
improves the quality of the generated images as well as the size and position of objects in the frame.
However, due to the absence of temporal motion information unique to videos, 
the objects struggle to adopt postures suitable for subsequent motion generation.
Consequently, the results with only image priors are not ideal as reference inputs for video generation.
In contrast, our Uni-C not only imposes constraints on image quality 
but also incorporates video diffusion priors, providing additional motion cues for the generated objects. 
As a result, the generated videos exhibit significantly improved dynamic degrees.

\section{Comparison with Prompt Engineering}
\label{sec:prompt}
Our Uni-C is designed to apply classifier guidance during the denoising sampling process, 
utilizing the priors of Video Diffusion models to generate results enriched with motion cues.
For instance, it ensures that generated objects are more appropriately positioned and postured for motion generation.
This suggests an alternative approach: directly incorporating human prior knowledge
through prompt engineering to constrain the completeness and posture of foreground objects.
To verify the effectiveness of prompt engineering, we modified the prompts of previous cases by incorporating human-curated text conditions 
to evaluate whether the SD model could produce the expected results without guidance.
As shown in Fig~\ref{fig:prompt}, additional text terms like "full body" and "dynamic pose" are added to the original prompt
to generate images with more complete objects and postures better suited for motion generation.
While detailed prompts show some improvement compared to the vanilla results,
the overall quality does not reach the level achieved by our Uni-C.
These results indicate the limitations of relying solely on CLIP embeddings for precise control over image details, 
especially for abstract concepts like "dynamic pose". 
Consequently, achieving the same level of control as Uni-C through prompt engineering alone proves to be challenging.

\section{Labeling with Different Video Diffusion}

\begin{figure*}[!ht]
    \centering
    \includegraphics[width=1.0\textwidth]{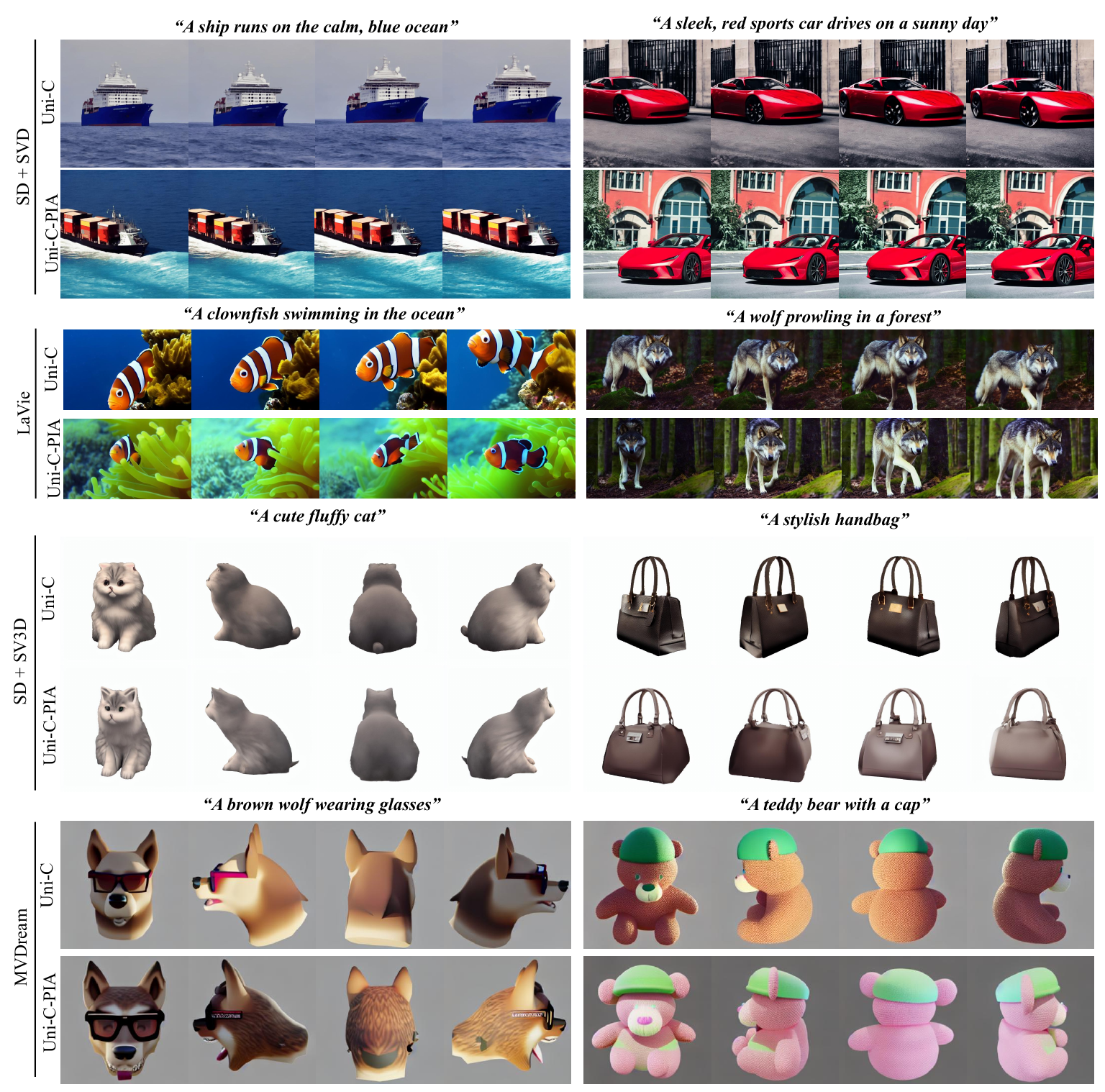}
    \vspace{-0.5em}
    \caption{\textbf{Visual comparison with different video diffusion models for constructing image-video pair data.} 
    Uni-C and Uni-C-PIA refer to classifiers trained with labels generated using SVD and PIA, respectively.
    The similar results exhibited by the two models demonstrate the commonality of video diffusion priors across different video generation models.
    }
    \vspace{-1em}
    \label{fig:pia_model}
\end{figure*}

To validate that the image-video pairs constructed for training Uni-C are equally effective
when applied to different video diffusion models, we replace SVD~\cite{blattmann2023stable} with PIA~\cite{zhang2024pia} in this section.
Building on images generated by the SD model, PIA performs the image-to-video process using text prompts.
Subsequently, GPT-4V is utilized to label the resulting image-video dataset. 
The classifier trained on labels generated through PIA is referred to as Uni-C-PIA.
Results are illustrated in Fig~\ref{fig:pia_model}, where we evaluate workflow-based video generation, 3D generation tasks, 
and individual video generation alongside 3D generation tasks. 

To ensure fairness in comparison, the same prompts as those used in the main text are adopted.
This setting allows us to analyze the differences between Uni-C and Uni-C-PIA
when the video diffusion model is switched from SVD to PIA.
As shown in Fig~\ref{fig:pia_model}, 
under identical prompt conditions, the results of Uni-C and Uni-C-PIA exhibit notable similarities, 
indicates that in our labeled data collection process,
GPT-4V can provide satisfactory labels for outputs generated by different Video Diffusion Models,
based on the designed rules.
This comparison also demonstrates that video diffusion priors exhibit
commonality across different video diffusion models, 
further validating the soundness of our Uni-C.

\section{Additional Qualitative Results}

\begin{figure*}[htp]
    \centering
    \includegraphics[width=1.0\textwidth]{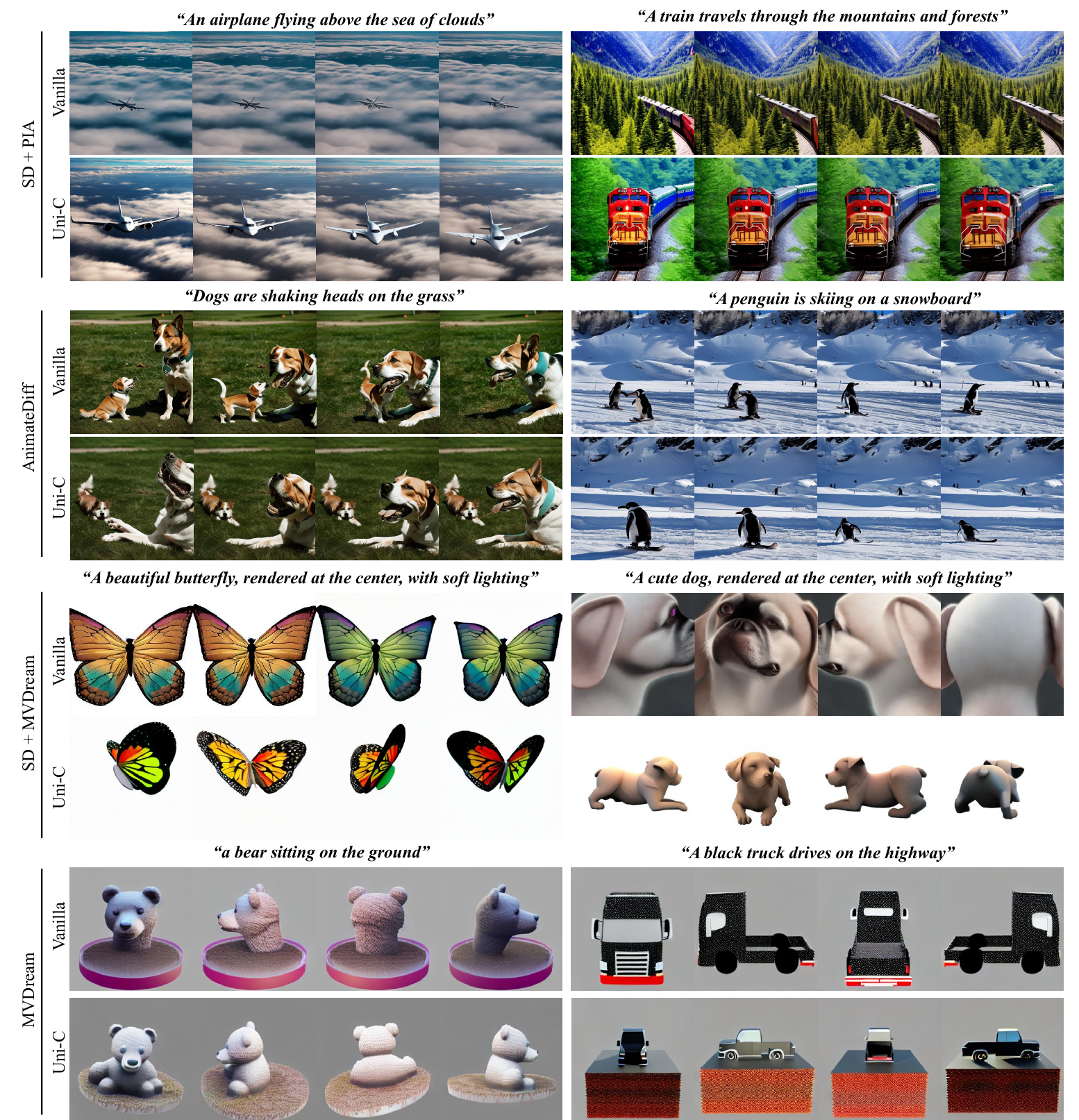}
    \caption{\textbf{Additional qualitative results on video and 3D generation tasks.} 
    }
    \label{fig:add_case}
\end{figure*}

This section presents additional qualitative results.
As illustrated in Fig.~\ref{fig:add_case}, we showcase of Uni-C applied to workflow-based video and 3D generation tasks, 
as well as individual text-to-video and text-to-3D tasks. These examples demonstrate
the performance of our Uni-C across various workflow-based and individual tasks.

\end{document}